# Remote Sensing Scene Classification with Masked Image Modeling (MIM)

Liya Wang[1], Alex Tien[2]
*The MITRE Corporation, McLean, VA, 22102, United States*

**Remote sensing scene classification has been extensively studied for its critical roles in geological survey, oil exploration, traffic management, earthquake prediction, wildfire monitoring, and intelligence monitoring. In the past, the Machine Learning (ML) methods for performing the task mainly used the backbones pretrained in the manner of supervised learning (SL). As Masked Image Modeling (MIM), a self-supervised learning (SSL) technique, has been shown as a better way for learning visual feature representation, it presents a new opportunity for improving ML performance on the scene classification task. This research aims to explore the potential of MIM pretrained backbones on four well-known classification datasets: Merced, AID, NWPU-RESISC45, and Optimal-31. Compared to the published benchmarks, we show that the MIM pretrained Vision Transformer (ViTs) backbones outperform other alternatives (up to 18% on top 1 accuracy) and that the MIM technique can learn better feature representation than the supervised learning counterparts (up to 5% on top 1 accuracy). Moreover, we show that the general-purpose MIM-pretrained ViTs can achieve competitive performance as the specially designed yet complicated Transformer for Remote Sensing (TRS) framework. Our experiment results also provide a performance baseline for future studies.**

## I. Introduction

In the past several years, remote sensing images have become easily accessible due to more and more devices dedicated to data collection. As artificial intelligence (AI) is booming, the methods for performing computer vision (CV) tasks on those images have advanced rapidly. One common CV task is remote sensing scene classification, which takes an image and correctly labels it to a predefined class. Scene classification is an important task for many applications such as land management, urban planning, wildfire monitoring, geological survey, oil exploration, traffic management, earthquake prediction, and intelligence monitoring [1].

The machine learning (ML) methods for remote sensing scene classification have been studied extensively (e.g., [2], [3], [4], [5],[6], [7], [8], [9], [10], [11]). Most studies in the past adopted the classical two-stage training paradigm: pre-training plus fine-tuning. See Figure 1 for illustration, where the backbones for feature extractions such as ResNet [12], Vision Transformer (ViT) [13], and Swin-T [14] are commonly pretrained in a supervised manner on ImageNet dataset [15], and then linear classification head layers are added on top of backbones and got fine-tuned on the task datasets in a supervised learning means, too.

Although ViTs have shown impressive performance over their convolution neural networks (CNNs) counterparts, they are prone to overfit the small datasets and usually require a large quantity of labeled datasets. In natural language processing (NLP), self-supervised pre-training methods like masked language modeling (MLM) have successfully addressed this problem. Motivated by MLM, BEiT [16] proposes Masked Image Modeling (MIM) to relieve the label-hungry problem of Transformers [17] while achieving impressive performance on various downstream tasks [18]. As such, the recent trend in CV has switched to adopting self-supervised learning (SSL) techniques (e.g., contrastive learning, MIM) for pre-training; see Figure 2 for illustration. SSL methods can pretrain backbones with unlabeled data by leveraging the structure present in the data itself to create supervised tasks (such tasks are often referred to as "pretext tasks").

To date, various MIM techniques for visual feature representation learning have been proposed (see Table 1 image and video rows for a comprehensive list). The most famous one is Masked Autoencoder (MAE) [19], which owns a very simple learning architecture but has been proven to be a strong and scalable pre-training framework for visual representation learning. MAE has attracted unprecedented attention and got various derivatives (e.g., CAE [20], ConvMAE [21], CMAE [22], GreenMAE [23], MixMIM [24]).

---

[1] Lead Artificial Intelligence Engineer, Department of Operational Performance
[2] Principal Engineer, Department of Operational Performance

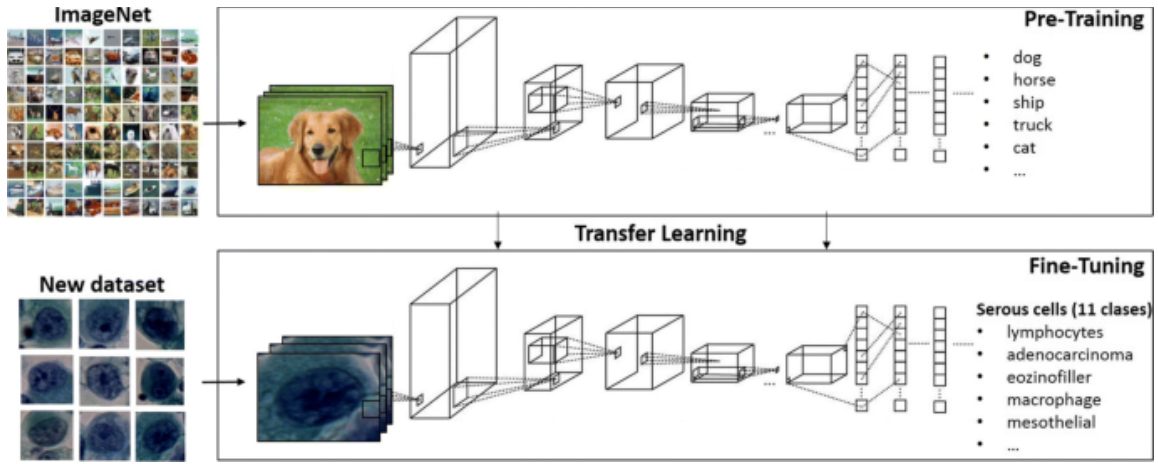

Figure 1. Pretraining in supervised manner plus fine-tuning [25].

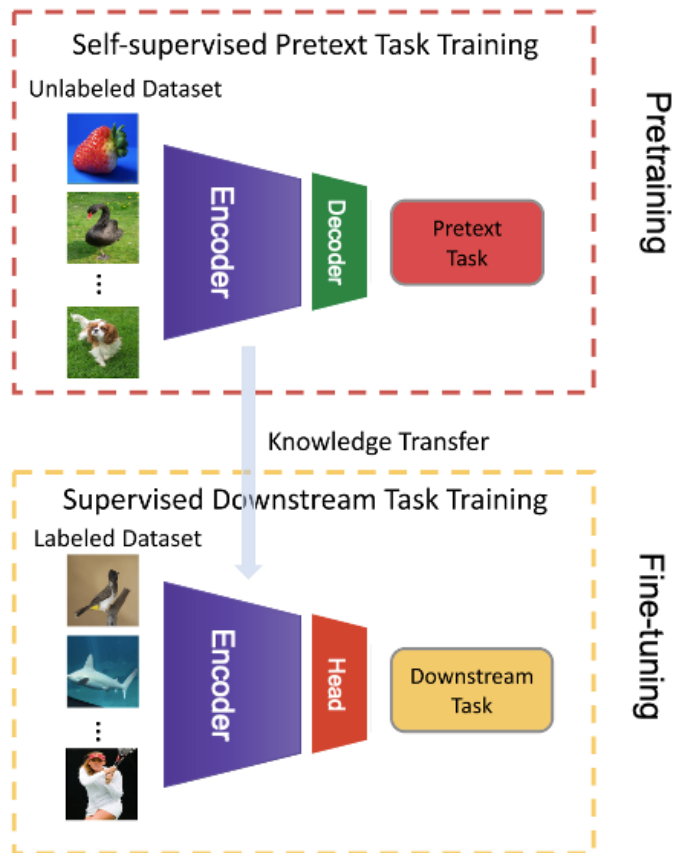

Figure 2. Pretraining in self-supervised manner plus fine-tuning paradigm [26].

To authors' knowledge, no research has ever explored MAE pretrained backbones for scene classification. Therefore, this research aims to evaluate MAE pretraining capability for the task. The remainder of the paper is organized as follows: Section II describes the related work, and Section III presents the selected four scene classification datasets. The results and discussion are presented in Section IV and V, respectively. Section VI is the conclusion.

Table 1. MIM techniques for visual feature learning

| Domain | Sub-Domain | Research Papers |
|---|---|---|
| Vision | Image | BEiT v1 [16], v2 [27], MAE [19], SimMIM [28], ADIOS [29], AMT [30], AttMask [31], Beyond-Masking [32], BootMAE [33], CAE [20], CAN [34], ConvMAE [21], Contrastive MAE [22], ContrastMask [35], dBOT [36], DMAE [37], Denoising MAE [38], GreenMAE [23], iBOT [39], LoMaR [40], LS-MAE [41], MaskAlign [42], MaskDistill [18], MaskFeat [43], MaskTune [44], MetaMask [45], MFM [46], MILAN [47], MixMask [48], MixMIM [24], MRA [49], MSN [50], MST [51], MultiMAE [52], MVP [53], RC-MAE [54], SDMAE [55], SemMAE [56], SdAE [57], SupMAE [58], U-MAE [59], UM-MAE [60] |
| | Video | AdaMAE [61], Bevt [62], MAM2 [63], MAR [64], MaskViT [65], M3Video [66], MCVD [67], MotionMAE [68], OmniMAE [69], Spatial-Temporal [70], SSVH [71], VideoMAE [72], Vimpac [73], VRL [74] |

## II. Related Work

**A. Vision Transformer (ViT)**

ViT [13] was proposed to make the standard Transformer [17] architecture process image data efficiently. Unlike traditional CNNs whose filters can only attend locally, the global attention mechanism of ViTs can integrate information across the entire image. ViTs outperform the CNNs by almost four times in terms of computational efficiency and accuracy [75], and are replacing CNNs in the CV field.

Although Transformer architectures have achieved so much success in the natural language processing (NLP) domain for a while, their success in the CV field was slow due to the different data characteristics between text and image (see Table 2 for comparison). An image could have thousands of pixels; in contrast, the input sequence length of text data is in tens. The computation complexity of Transformer is $O(n^2d)$, where $n$ and $d$ are the input sequence length and embedding length, respectively.

Table 2. Data characteristics comparison

| Text | Image |
|---|---|
| 1-dimensional | 2-dimensional |
| Discrete | Continuous |
| Low redundancy | High redundancy |
| Low computation cost due to small $n$ | High computation cost due to large $n$ |

To deal with the problem, ViTs adopt a special method to preprocess the image data, which can be described as follows (see Figure 3 for illustration):

Step 1. Split an image into non-overlapping patches (fixed sizes, e.g., 16 × 16 or 32 × 32).
Step 2. Flatten the image patches.
Step 3. Encode the flattened patches into linear embeddings.
Step 4. Add positional embeddings to the patch embeddings of Step 3.
Step 5. Feed the sequence as an input to Transformer encoder.

This way, they can reduce input sequence length to $n' = \frac{W \times H}{p^2}$, where $W, H,$ and $p$ are width, height, and patch size of the image, respectively. With such preprocessing, Transformer architecture can process image data much efficiently. Next, the relevant MIM methods tested in our work will be presented.

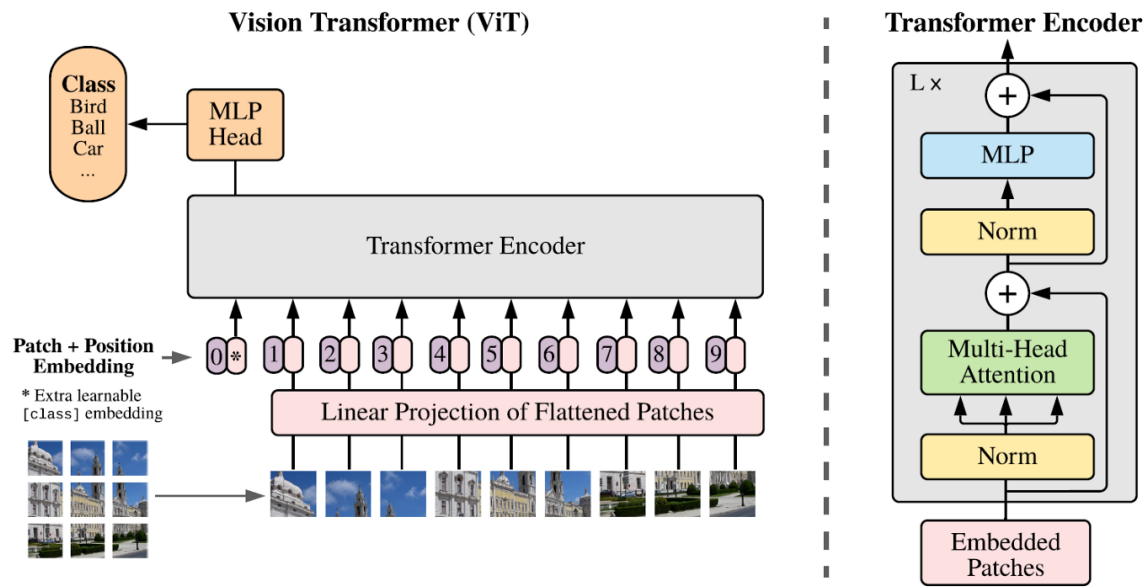

**Figure 3. ViT architecture [76].**

## B. Masked Autoencoder (MAE)

MAE is an asymmetric autoencoder that uses ViTs in both its encoder and decoder, and the size of the decoder is smaller than the encoder, as illustrated in Figure 4. It directly infers masked patches from the unmasked ones with a simple loss of mean squared error (MSE). To save computation, the encoder only works on the unmasked patches; in contrast, the decoder works on both masked and unmasked patches trying to predict the original images. The masking ratio can be set up to 75%, which is considerably higher than that in BERT (typically 15%) [77] or earlier MIM methods (20% to 50%) [16], [78]. MAE's ablation study also points out that a high masking ratio is good for fine-tuning and linear probing [19]. With those meticulous designs, MAE is three times (or more) faster than BEiT [16] while achieving superior performance [19].

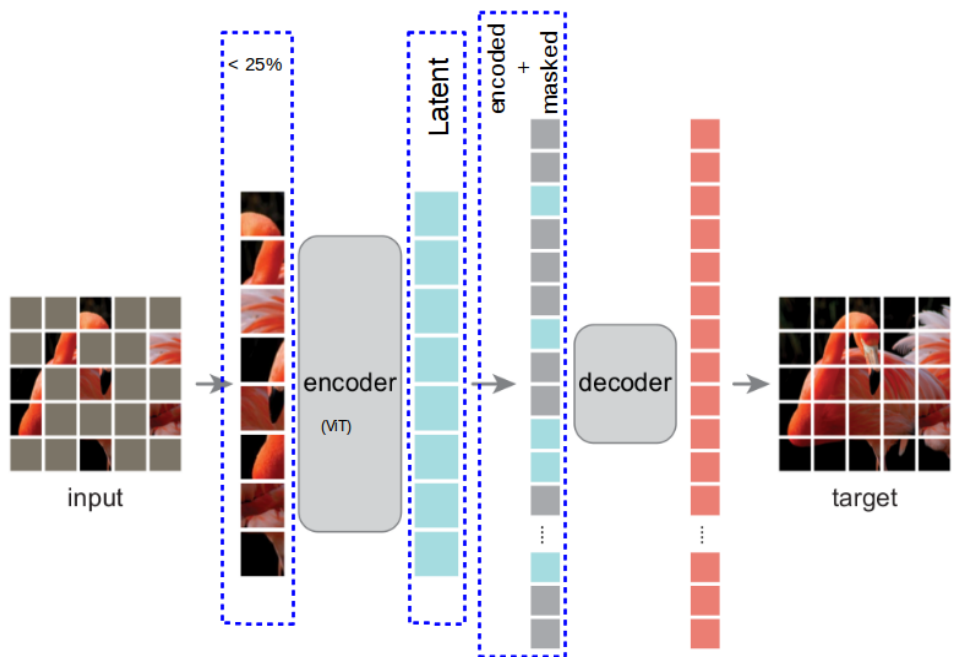

**Figure 4 MAE architecture [19].**

## C. Context autoencoder (CAE)

Context autoencoder (CAE) [20] was also proposed for self-supervised representation pre-training of ViTs. Unlike MAE, the pretext goal of CAE is to predict the masked patches from the visible patches instead of the whole image. The architecture of CAE consists of an encoder, a latent contextual regressor with an alignment constraint, and a decoder (see Figure 5 for illustration). The working pipeline of CAE is as follows:

Step 1. The visible patches are fed into the encoder to get their representations.

Step 2. The encoded representations of visible patches together with mask queries are then fed to the contextual regressor to get the representation of masked patches. It should be noted that masked queries are learnable during the training.

Step 3. The masked patches' presentations are also computed from the encoder.

Step 4. An alignment constraint is applied on the outputs of Step 2 and Step 3, which are expected to be the same in representation space, to calculate a loss value.

Step 5. Step 2's results are fed to the decoder for generating the masked tokens, which are then compared to the targets generated by feeding masked patches to the pretrained DALL-E tokenizer [79]. The difference here formulates another loss value.

Step 6. Combine losses in Step 4 and Step 5 together for the optimization.

Compared to BEiT [16], which combines the encoding and pretext task completion roles together, CAE separates them. This way, it can improve the representation learning capacity, which further supports downstream tasks. The masking ratio in CAE is 50%, which is lower than 75% of MAE.

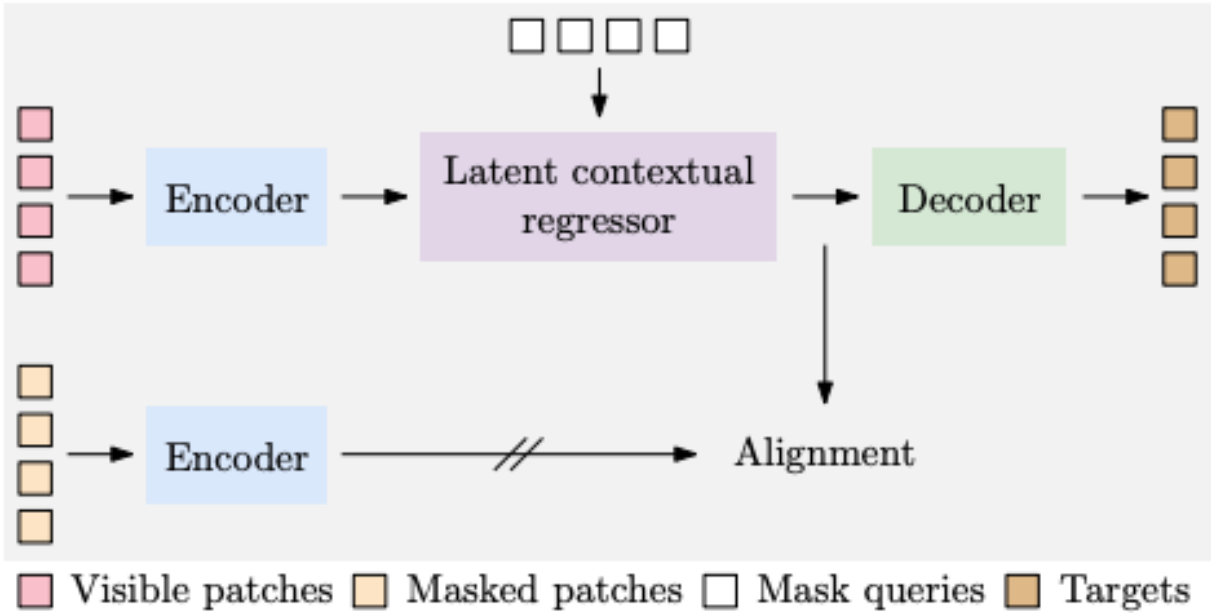

**Figure 5. CAE architecture [20].**

## D. Masked Convolution Meets Masked Autoencoders (ConvMAE)

ConvMAE [21], a derivative of the popular MAE [19], is proposed to train scalable visual representation with hybrid convolution-transformer architectures and masking convolution. It integrates both merits of local inductive bias from CNNs and global attention of ViTs. Although the modifications to the original MAE are minimal, ConvMAE has demonstrated great success on pre-training visual representations for improving the performance of various tasks [21]. ConvMAE can also provide multi-scale features while avoiding the discrepancy between pre-training and fine-tuning.

Like MAE, ConvMAE architecture still consists of two parts: encoder and decoder (see Figure 6). However, its encoder is a hybrid convolution-transformer architecture, and its decoder part is still made of ViT. In addition, ConvMAE introduces a hierarchical masking strategy together with masked convolution to make sure that only a small number of visible tokens are fed into the transformer encoder layers (see Figure 6, top row). As shown in Figure 6, the encoder has three stages with output spatial resolutions of $\frac{W}{4} \times \frac{H}{4}$, $\frac{W}{8} \times \frac{H}{8}$, and $\frac{W}{16} \times \frac{H}{16}$, respectively, where $H$ and $W$ are the height and width of the input image. The encoder can generate multi-scale features $E_1$, $E_2$, and $E_3$, which capture both fine- and coarse-grained image information. The transformer blocks of encoder in Stage 3 aggregate and fuse three features together (see the bottom row blue block in Figure 6 for illustration) and send them to the decoder of ConvMAE, which still works on both visible and masked tokens (see the middle row green block in Figure 6 for illustration). The loss function is the same as the one used in MAE in which only masked patches are considered for the loss values calculation. Next, we will present the image datasets selected by this research for evaluating the performance of various MIM pretrained backbones on remote sensing scene classification.

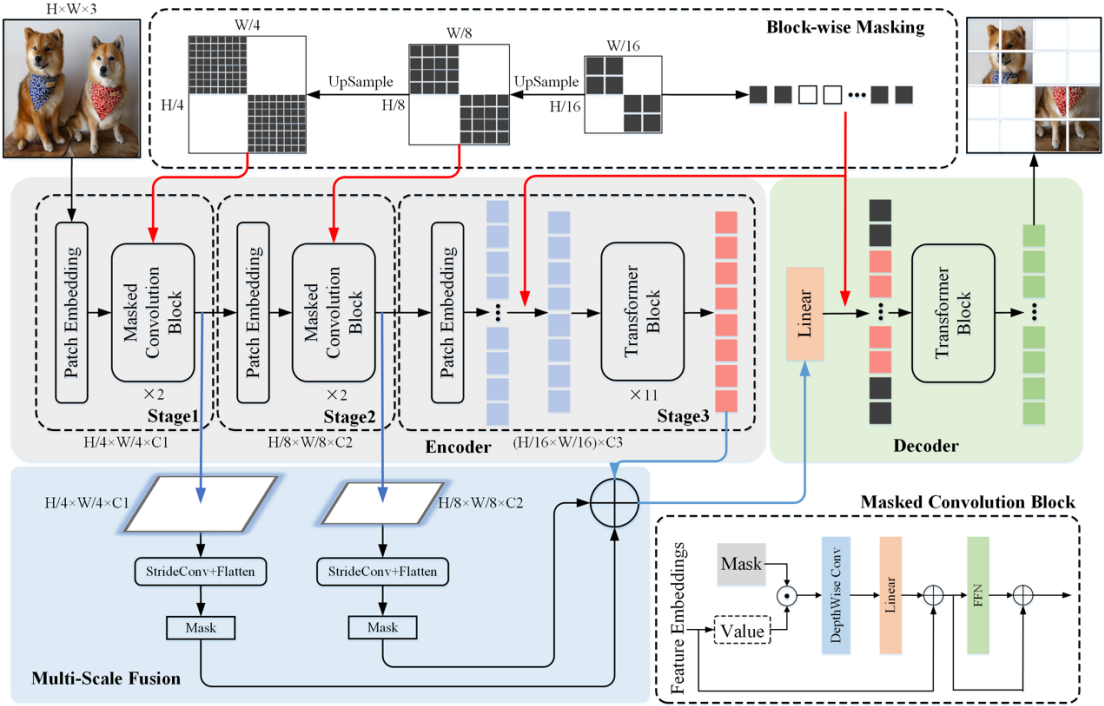

**Figure 6. ConvMAE architecture [21].**

### III. Datasets

We have chosen four well-known remote sensing scene image classification datasets for evaluation: 1) Merced land-use dataset [80], 2) Aerial image dataset (AID) [81], 3) NWPU-RESISC45 [2], and 4) Optimal-31 dataset [82]. The characteristics of these four datasets are summarized in Table 3. The rest of this section provides a short introduction for each of the datasets.

**Table 3. Selected classification dataset information**

| Datasets | Images per class | Scene class | Total images | Spatial resolution(m) | Image sizes | Year |
|---|---|---|---|---|---|---|
| UC Merced Land Use | 100 | 21 | 2,100 | 0.3 | 256×256 | 2010 |
| AID | 220~420 | 30 | 10,000 | 0.5~8 | 600×600 | 2017 |
| NWPU-RESISC45 | 700 | 45 | 31,500 | 0.2-30 | 256×256 | 2017 |
| OPTIMAL-31 | 60 | 31 | 1,860 | 0.3 | 256×256 | 2017 |

#### A. Merced Dataset

Merced Dataset [80] was released in 2010, and has 2,100 RGB images of 21 land-use scene classes. Each class contains 100 images of size 256 × 256 pixels with 0.3 m resolution. The images were extracted from the United States Geological Survey National Map [83]. Figure 7 shows the image samples from the 21 classes.

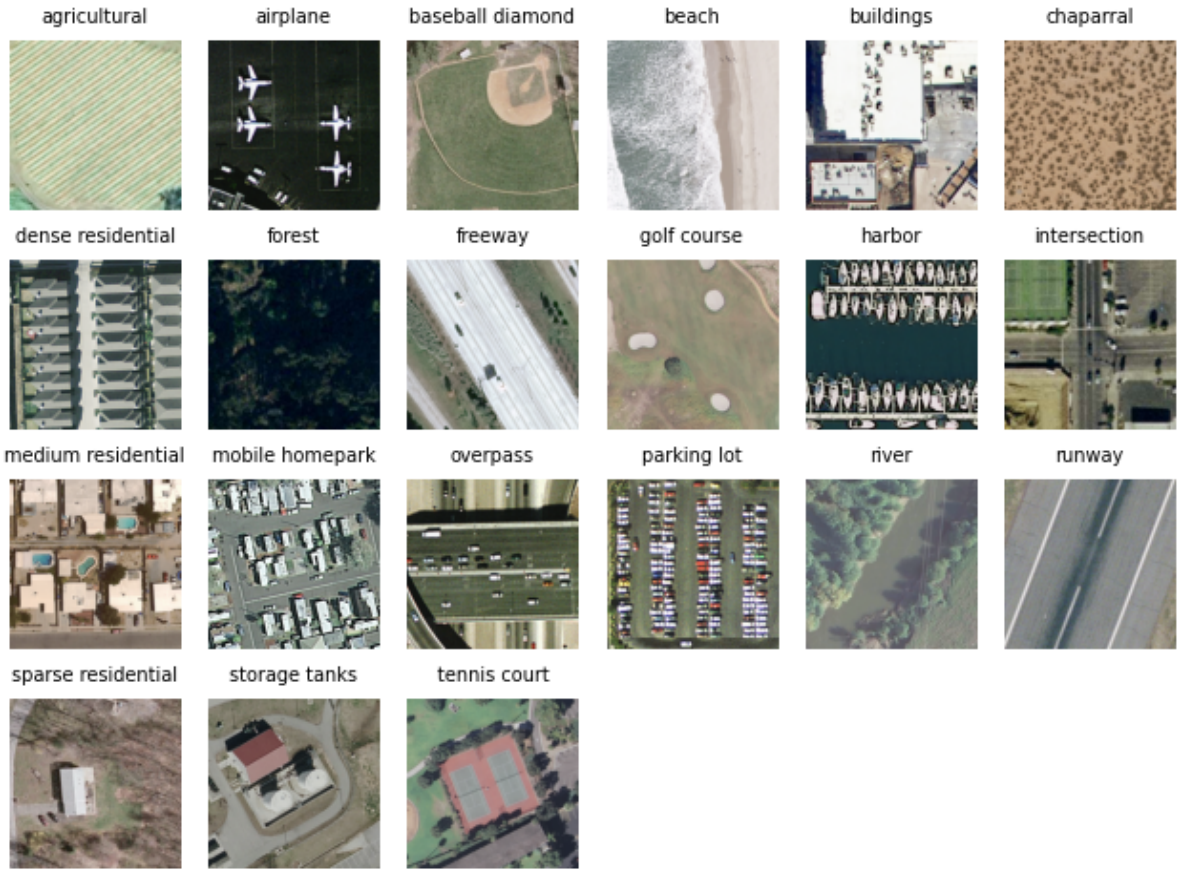

**Figure 7. UC Merced example images.**

**B. Aerial image dataset (AID) Dataset**

AID [81] dataset was published in 2017 by Wuhan University, China. It has 10,000 images. The images are classified into 30 classes with 220 to 420 images per class. The images were cropped from Google Earth imagery measuring 600 × 600 pixels with a resolution varying from 0.5 m to about 8 m. Figure 8 shows the image samples from the 30 classes.

**C. NWPU-RESISC45 Dataset**

The NWPU-RESISC45 [2] dataset was published in 2017 by Northwestern Polytechnical University, China. It contains 31,500 remote sensing images grouped into 45 scene classes. Each class includes 700 images with a size of 256 × 256 pixels, and the spatial resolution varies from about 30 to 0.2 m per pixel for most of the scene classes, except for the classes of island, lake, mountain, and beach, which have lower spatial resolutions. This dataset was also extracted from Google Earth that maps Earth by the superimposition of images obtained from satellite imagery, aerial photography, and geographic information system (GIS) onto a 3-D globe. The 31,500 images are collected from more than 100 countries and regions across the world, including developing, transition, and highly developed economies. Figure 9 shows one sample of each class from this dataset.

**D. Optimal-31 Dataset**

Optimal-31 [82] dataset was created in 2017 by Northwestern Polytechnical University, China. It contains 31 scene classes. Each class consists of 60 images with size of 256 × 256 pixels. Figure 10 shows an example image for every class. The pixel resolution for the images is 0.3 m.

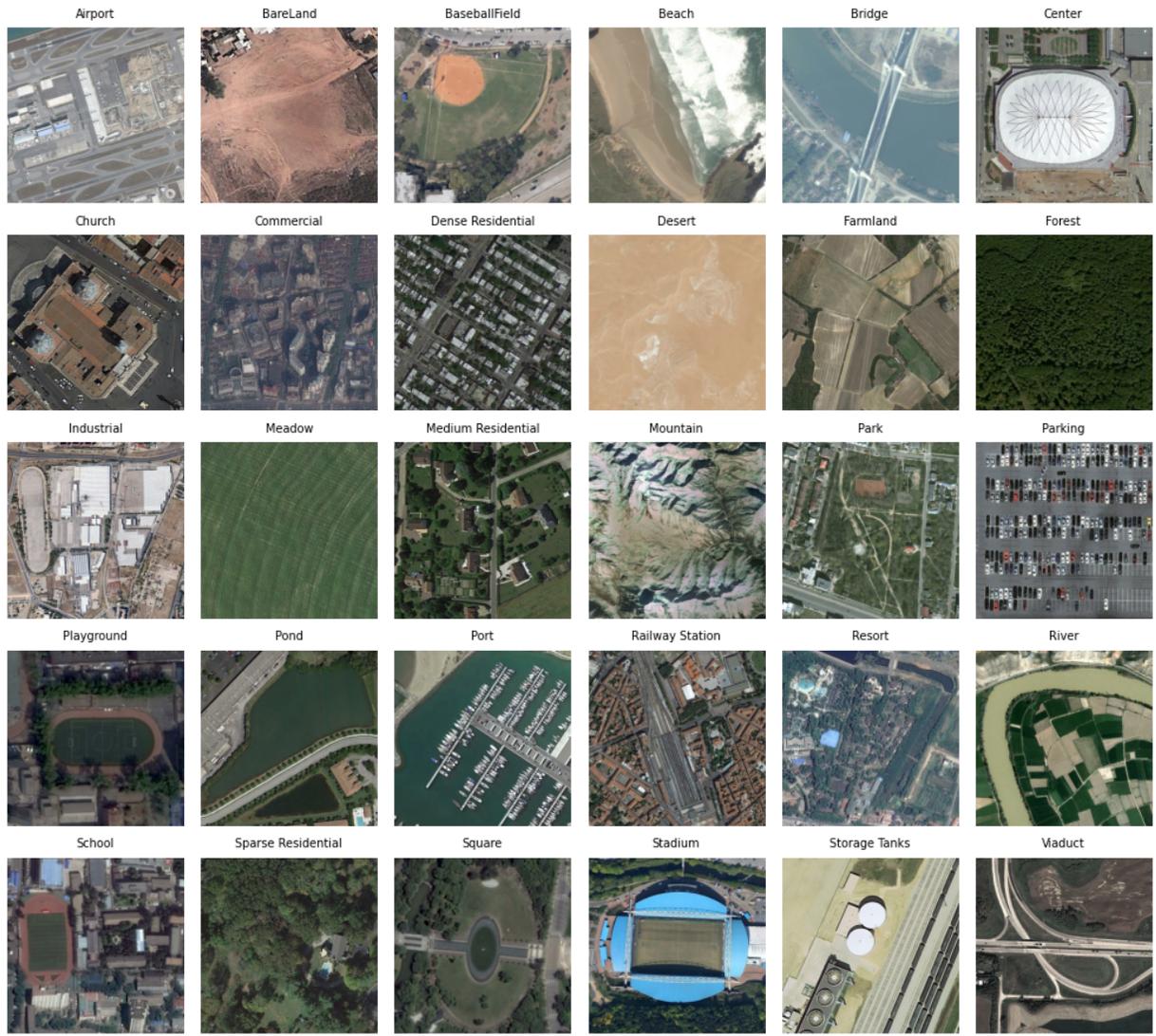

**Figure 8. AID example images.**

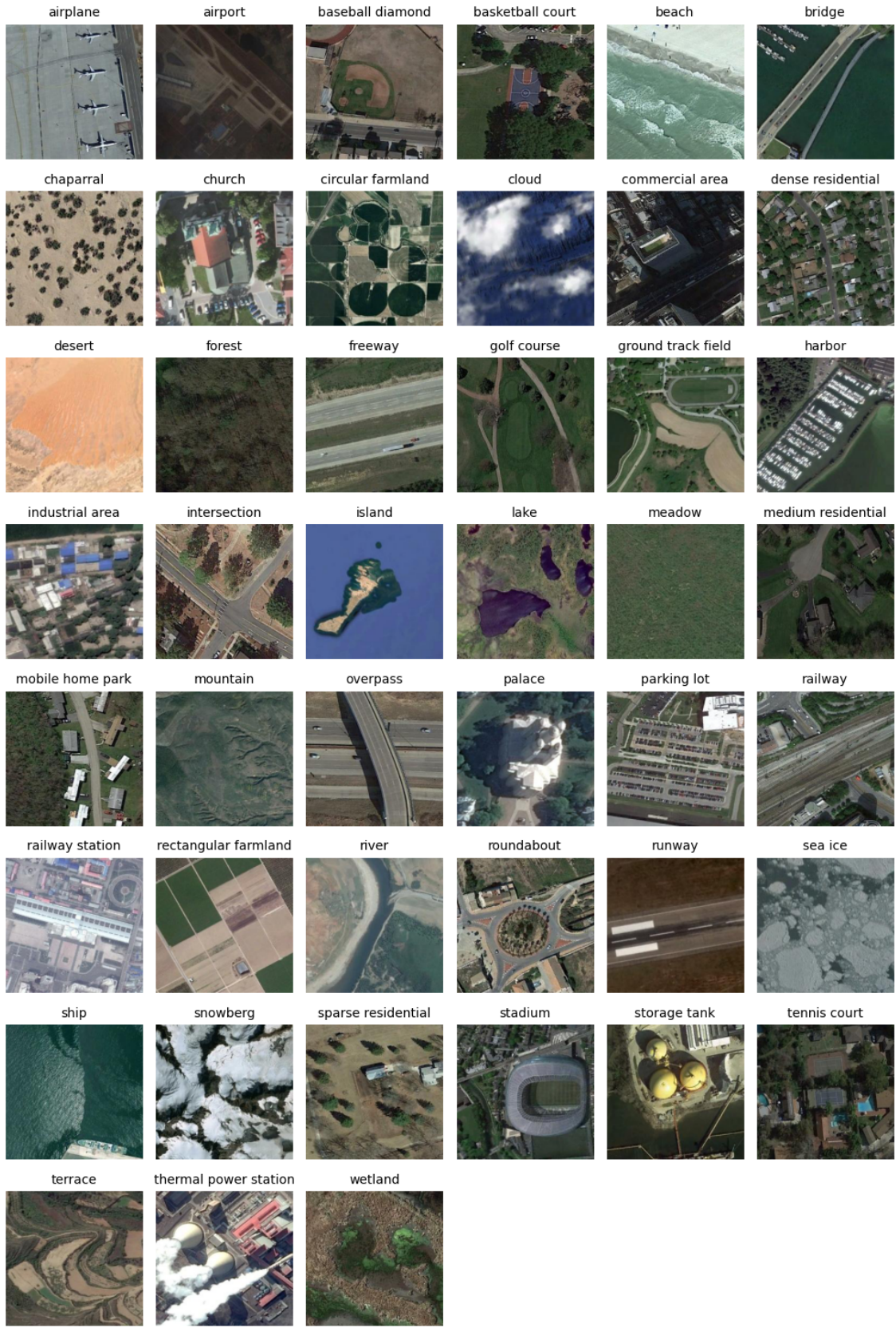

**Figure 9.** NWPU-RESISC45 example images.

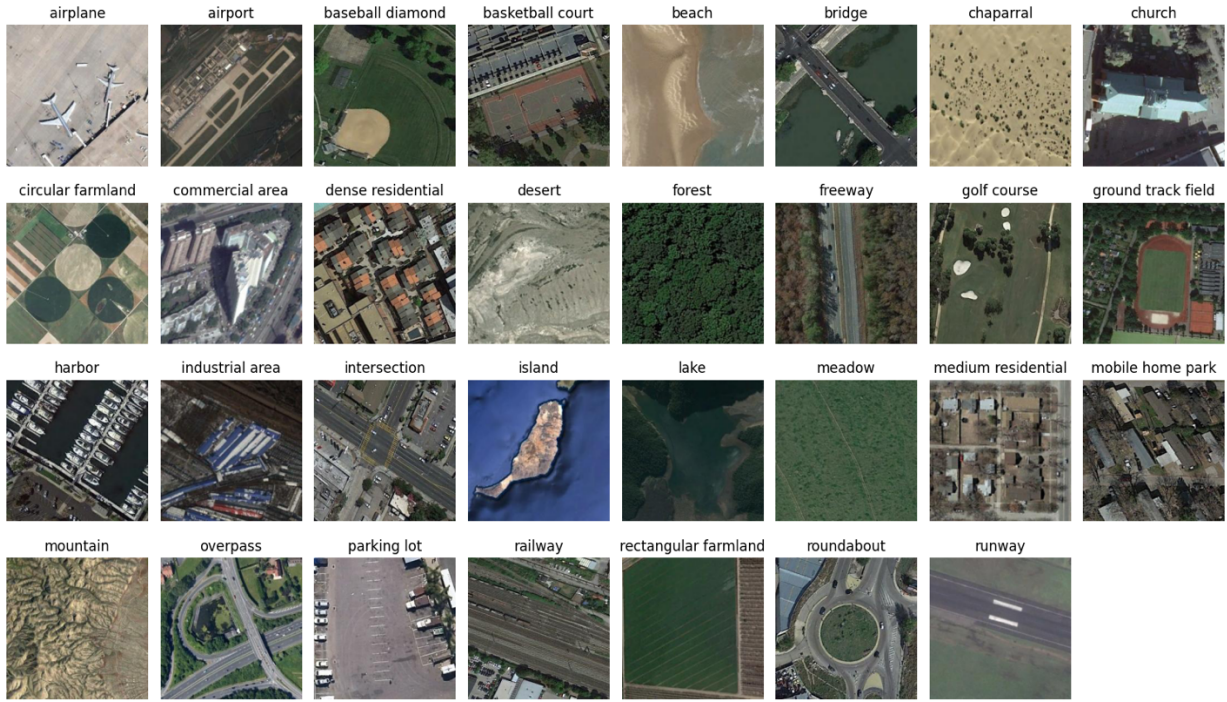

Figure 10. Optimal-31 example images.

## IV. Results

This section presents the four experiment results of remote sensing scene image classification with backbones pretrained with MAE, CAE, and ConvMAE, respectively. The results are also compared with those from 17 algorithms listed in [5], which contains the results from 16 CNNs and one specially designed Transformer-based architecture, Transformers for Remote Sensing (TRS). According to [5], TRS has achieved state-of-the-art performance. The implementation details and the corresponding results are presented as follows.

### A. Experimental Setup

For a fair comparison, we tried to follow the same experiment setup laid out in [5] if possible. The training equipment setup is shown in Table 4. First, we downloaded the pretrained backbones directly from their official GitHub websites. Then, we carried out fine-tuning on the tested datasets. In specifics, all experiments were fine-tuned for 80 epochs. Optimizer was Adam. The initial learning rate was set to 0.0004, and weight decay was set to 0.00001. All images were reshaped to 224 × 224 sizes, and the batch size was set to 16. The top 1 accuracy (acc1) was used for the evaluation. The best performance metrics are highlighted in bold in the result tables.

Table 4. Experimental environment

| Operation System | Linux |
|---|---|
| CPU | 2xAMD EPYC 7262 8-Core Processor |
| Memory | 250 GB |
| Framework | PyTorch 1.13.1 |
| GPUs | 4xA100 |

### B. Data Augmentation Strategies

During the fine-tuning stage, we adopted data augmentation for better performance. We adopted the MixUp [84] and CutMix [85] data augmentation techniques. For MixUp, two images are merged by linearly interpolating them along with their class labels to create a new training instance. CutMix is to cut a patch of one image and replace it with a patch from another image in the dataset (see Figure 11 for examples). We set the parameters as 0.8 and 1.0 for MixUp and CutMix, respectively.

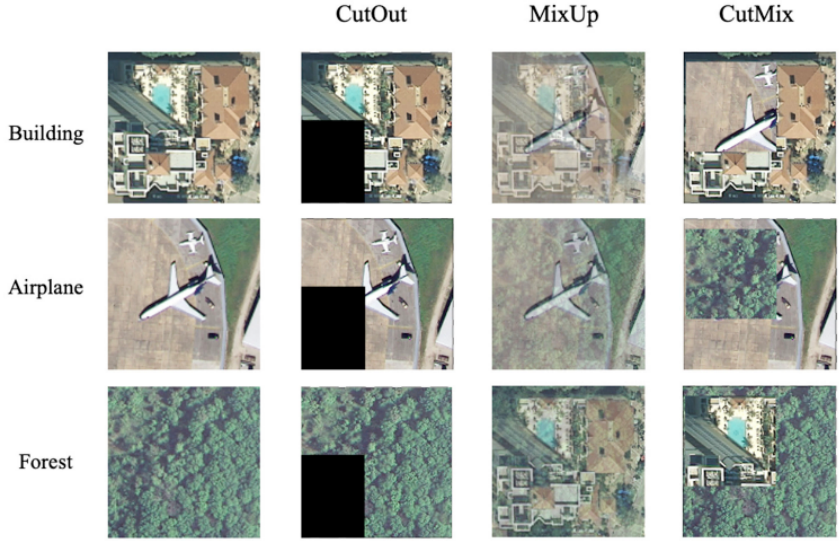

Figure 11. Examples of applying data augmentation techniques on Merced dataset [1].

C. Merced Dataset Classification Results

For this dataset, 80% of the images were used as the training dataset, and 20% as the testing dataset. It should be noted that for column names, PT represents pre-training; FT is fine-tuning; lr is learning rate; and acc1 is top 1 accuracy rate. The results are listed in Table 5, from which we can see that large backbones ViT-L and ViT-H can achieve 100% accuracy. Compared to the previously published results, ranging 94.31% to 99.52%, from 17 deep learning methods listed in Table 3 of [5], no one has achieved such good performance. In addition, all MIM methods but ConvMAE perform better (99.76% to 100%) than the TRS method (99.52%) [5], which is the best method listed in Table 3 of [5].

Table 5. Classification accuracy on UC-Merced dataset (80% for training)

| Method | Pretraining | Backbone | PT-Supervision | Masked-Perc | FT-Epochs | lr | FT-acc1(%) | Source |
|---|---|---|---|---|---|---|---|---|
| MAE | 1k, MAE | ViT-B | RGB | 75% | 80 | 0.0004 | 99.76% | ours |
| MAE | 1k, MAE | ViT-L | RGB | 75% | 80 | 0.0004 | **100.00%** | ours |
| MAE | 1k, MAE | ViT-H | RGB | 75% | 80 | 0.0004 | **100.00%** | ours |
| CAE | 1k, CAE | ViT-B | DALLE | 50% | 80 | 0.0004 | 99.76% | ours |
| CAE | 1k, CAE | ViT-L | DALLE | 50% | 80 | 0.0004 | **100.00%** | ours |
| ConvMAE | 1k, ConvMAE | ConvViT-B | RGB | 75% | 80 | 0.0004 | 99.29% | ours |
| TRS | 1k,sup | - | Labels | None | 80 | 0.0004 | 99.52% | [5] |

D. AID Dataset Classification Results

For this dataset, 50% of the images were used as the training dataset, and 50% as the testing dataset. Table 6 lists the classification results for the AID dataset. Compared to the previously published results, ranging 86.39% to 98.48%, from 17 deep learning methods listed in Table 4 of [5], MIM methods achieve acc1 ranging from 97.5% to 98.15%, which still beats most of CNNs. It should be noted that our AID images were resized to 224 × 224 for using MIM pretrained backbones; instead, the TRS method used 600 × 600 image size, which could contribute to performance differences.

Table 6. Classification accuracy on AID dataset (50% for training)

| Method | Pretraining | Backbone | PT-Supervision | Masked-Perc | FT-Epochs | lr | FT-acc1(%) | Source |
|---|---|---|---|---|---|---|---|---|
| MAE | 1k, MAE | ViT-B | RGB | 75% | 80 | 0.0004 | 98.00% | ours |
| MAE | 1k, MAE | ViT-L | RGB | 75% | 80 | 0.0004 | 97.90% | ours |
| MAE | 1k, MAE | ViT-H | RGB | 75% | 80 | 0.0004 | 98.14% | ours |
| CAE | 1k, CAE | ViT-B | DALLE | 50% | 80 | 0.0004 | 97.50% | ours |
| CAE | 1k, CAE | ViT-L | DALLE | 50% | 80 | 0.0004 | 97.82% | ours |
| ConvMAE | 1k, ConvMAE | ConvViT-B | RGB | 75% | 80 | 0.0004 | 97.92% | ours |
| TRS | 1k,sup | - | Labels | None | 80 | 0.0004 | **98.48%** | [5] |

## E. NWPU-RESISC45 Dataset Classification Results

For this dataset, 20% of the images were used as the training dataset, and 80% as the testing dataset. Table 7 lists the classification results for the NWPU-RESISC45 dataset. Compared to the previously published results (76.85% to 95.56%) from 17 deep learning methods listed in Table 5 of [5], the MAE with ViT-H backbone (95.61%) can beat the previous best TRS method (95.56%). Once again, the experiment demonstrates the MIM pretrained backbones perform better (94.40% to 95.61%) than most of the CNNs, whose performances range from 76.85% to 94.43%.

Table 7. Classification accuracy on NWPU-RESISC45 dataset (20% training)

| Method | Pretraining | Backbone | PT-Supervision | Masked-Perc | FT-Epochs | lr | FT-acc1(%) | Source |
|---|---|---|---|---|---|---|---|---|
| MAE | 1k, MAE | ViT-B | RGB | 75% | 80 | 0.0004 | 94.40% | ours |
| MAE | 1k, MAE | ViT-L | RGB | 75% | 80 | 0.0004 | 95.31% | ours |
| MAE | 1k, MAE | ViT-H | RGB | 75% | 80 | 0.0004 | **95.61%** | ours |
| CAE | 1k, CAE | ViT-B | DALLE | 50% | 80 | 0.0004 | 94.71% | ours |
| CAE | 1k, CAE | ViT-L | DALLE | 50% | 80 | 0.0004 | 95.45% | ours |
| ConvMAE | 1k, ConvMAE | ConvViT-B | RGB | 75% | 80 | 0.0004 | 95.17% | ours |
| TRS | 1k,sup | - | Labels | None | 80 | 0.0004 | 95.56% | [5] |

## F. OPTIMAL-31 Dataset Classification Results

For this dataset, 80% of the images were used as the training dataset, and 20% as the testing dataset. Table 8 lists the classification results for this dataset. Compared to the previously published results (81.22% to 95.97%) from 10 deep learning methods listed in Table 6 of [5], ConvMAE with ViT-B backbone (96.51%) can beat the best TRS method (95.97%). Once again, the experiment demonstrates the MIM pretrained backbones perform better (93.20% to 96.51%) than most of the CNNs (81.22% to 94.51%).

Table 8. Classification accuracy on Optimal31 dataset (80% training)

| Method | Pretraining | Backbone | PT-Supervision | Masked-Perc | FT-Epochs | lr | FT-acc1(%) | Source |
|---|---|---|---|---|---|---|---|---|
| MAE | 1k, MAE | ViT-B | RGB | 75% | 80 | 0.0004 | 93.20% | ours |
| MAE | 1k, MAE | ViT-L | RGB | 75% | 80 | 0.0004 | 95.70% | ours |
| MAE | 1k, MAE | ViT-H | RGB | 75% | 80 | 0.0004 | 95.70% | ours |
| CAE | 1k, CAE | ViT-B | DALLE | 50% | 80 | 0.0004 | 95.70% | ours |
| CAE | 1k, CAE | ViT-L | DALLE | 50% | 80 | 0.0004 | 96.24% | ours |
| ConvMAE | 1k, ConvMAE | ConvViT-B | RGB | 75% | 80 | 0.0004 | **96.51%** | ours |
| TRS | 1k,sup | - | Labels | None | 80 | 0.0004 | 95.97% | [5] |

In addition, we compared the results between MIM and supervised learning pretrained ViTs listed in Table 8 and 9 of [5]. Obviously, for same backbone, our tested MIM methods learn much better representations than supervised pretraining methods (up to 5% on top 1 accuracy). For example, according to Table 8 of [5], supervised learning pretrained ViT-Base achieves 95.81% top-1 accuracy for Merced dataset (80% of training), and our tested MAE pretrained ViT-Base can achieve 99.76% top-1 accuracy, which denotes about 4% of improvement.

## V. Discussion

### A. Compare with Supervised Pretraining Methods

In addition, we compared the results of ViTs pretrained from our tested MIM methods and supervised learning methods published in the literature by far. Table 9 and Table 10 compares results from ViT-B and ViT-L which are pretrained with different methods, respectively. The best performance metrics are highlighted in bold in the result tables. Obviously, for same backbone, our tested MIM methods learn much better representations than supervised pretraining methods (up to 6% on top 1 accuracy). For example, supervised learning pretrained ViT-Base achieves 95.81% top-1 accuracy for Merced dataset (80% of training), and our tested MAE pretrained ViT-Base can achieve 99.76% top-1 accuracy, which denotes about 4% of improvement. In addition, MIM pretrained backbones with less data (1k) can outperform supervised learning methods with 21k data (see Table 9).

### B. Application Scenarios

MIM methods have been proved to be a great way of learning visual feature representation and can support multiple domains such as object detection, segmentation, multi-modal learning, reinforcement learning, time series, point cloud, 3D-mesh, and audio. Figure 12 summarizes the various applications of MIM published in the literature.

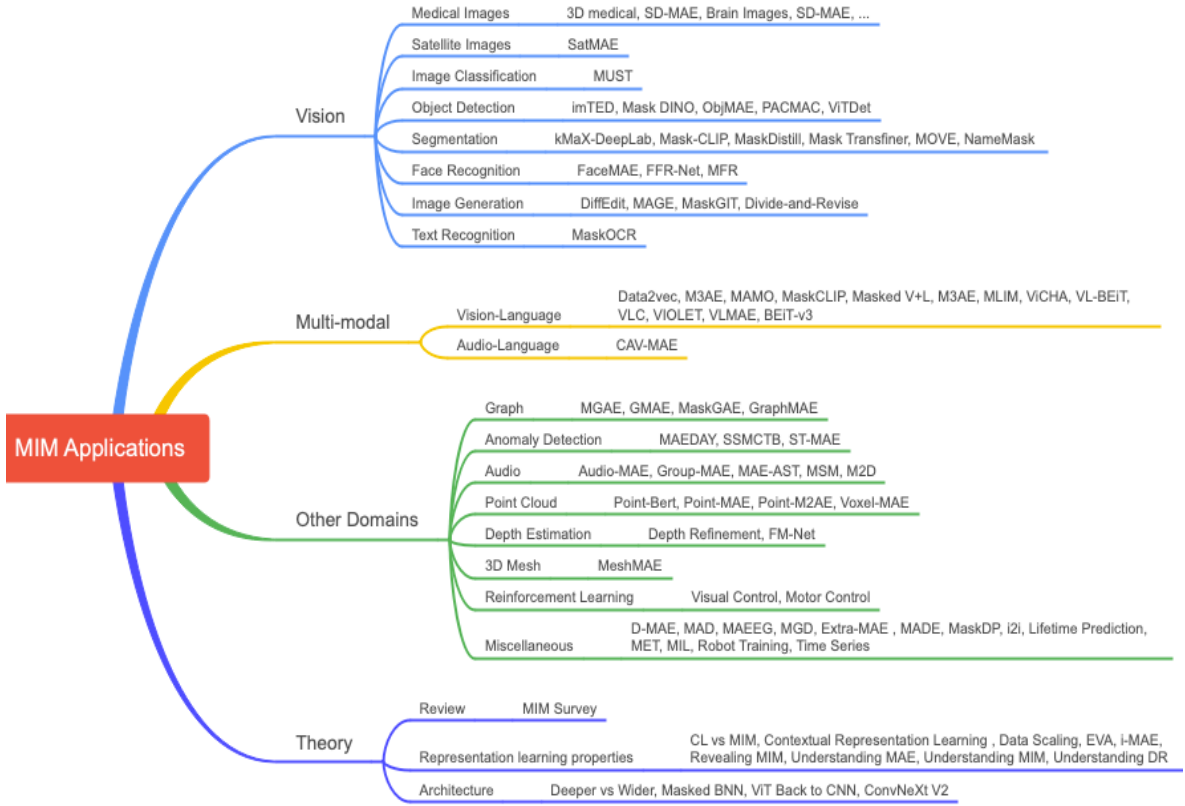

**Figure 12. MIM applications.**

**Table 9. Comparing results from ViT-B from different pretraining methods**

| Pretraining | Backbone | Merced (80% Training) | AID (50% Training) | NWPU-RESISC45 (20% Training) | Optimal-31 (80% Training) |
|---|---|---|---|---|---|
| 1k, sup [5] | ViT-B | 95.81% | 94.44% | 90.87% | 89.73% |
| 21k, sup [1] | ViT-B | 98.14% | 94.97% | 92.6% | 95.07% |
| 1k, MAE (ours) | ViT-B | **99.76%** | **98.00%** | 94.40% | 93.20% |
| 1k, CAE (ours) | ViT-B | **99.76%** | 97.50% | **94.71%** | **95.70%** |

**Table 10. Comparing results from ViT-L from different pretraining methods**

| Pretraining | Backbone | Merced (80% Training) | AID (50% Training) | NWPU-RESISC45 (20% Training) | Optimal-31 (80% Training) |
|---|---|---|---|---|---|
| 1k, sup [5] | ViT-L | 96.06% | 95.13% | 91.94% | 91.14% |
| 1k, MAE (ours) | ViT-L | **100%** | **97.90%** | 95.31% | 95.70% |
| 1k, CAE (ours) | ViT-L | **100%** | 97.82% | **95.45%** | **96.24%** |

## VI. Conclusion

This study has explored the use of the backbones pretrained by the newly proposed MIM methods (i.e., MAE, CAE, ConvMAE) to perform challenging remote sensing scene classification tasks. We carried out experiments on four well-known scene classification datasets: Merced, AID, NWPU-RESISC45, and Optimal-31. Our experiments demonstrated that MIM pretrained ViT backbones consistently beat CNN backbones (up to 18% on top 1 accuracy). In addition, for the same ViT backbone, MIM can learn better representation than the supervised learning counterparts (up to 5% on top 1 accuracy). Furthermore, our tested MIM methods can achieve on-par performance as specially designed yet complicated TRS architecture. Our experiment results also provided a performance baseline for future studies.

## Acknowledgments

The authors thank Dr. Kris Rosfjord and Dr. Heath Farris for their generous support of this project. We would also like to thank Mike Robinson, Bill Bateman, Lixia Song, Erik Vargo, and Paul A Diffenderfer of the MITRE Corporation for their valuable discussions, insights, and encouragement.

## NOTICE